\documentclass{article}

\usepackage[final]{corl_2026} %

\usepackage{multirow} 
\usepackage{graphicx}
\usepackage{amsmath}
\usepackage{amssymb}
\usepackage{subcaption}
\usepackage{makecell}
\usepackage{verbatim}
\usepackage{pifont}
\usepackage{xurl}
\usepackage{booktabs} 
\usepackage[table]{xcolor}
\usepackage{enumitem}

\title{Lights, Camera, Malfunction: When Illumination Robustness Leaves VLA Models Blind to Color}

\author{
  Marino Watanabe\qquad Takami Sato\qquad Kentaro Yoshioka \\
  Keio University \\
}

\begin{document}
\maketitle

\begin{abstract}
Vision-Language-Action (VLA) models have emerged as a powerful paradigm for general-purpose robot manipulation; however, their transition to real-world environments reveals vulnerabilities to minor environmental perturbations. We propose FLARE, an optimization framework that exploits these vulnerabilities via targeted illuminations, causing critical disruption of baseline tasks. While adversarial training is the standard countermeasure, we identify a critical and previously underestimated defensive pitfall: naive data augmentations condition VLA models to largely discard color as noise, biasing their visual perception toward shape cues. We expose this degradation through a diagnostic grayscale evaluation, in which a naively defended model maintains high success rates on grayscale inputs, while its success rate on benign, color-dependent real-world tasks drops to at most 48.8\%, compared with 78.8\% for the undefended SmolVLA baseline.
To address this, we propose ChromaGuard, a chroma-preserving adversarial training method. On a physical 6-DoF robotic platform using SmolVLA, we demonstrate that ChromaGuard achieves 95.0\% and 88.8\% success rates in benign and attacked color-dependent tasks, respectively. Demo videos are available at \url{https://keio-csg.github.io/ChromaGuard/}.
\end{abstract}


\section{Introduction}

\begin{figure}[t]
    \centering
    \includegraphics[width=\linewidth]{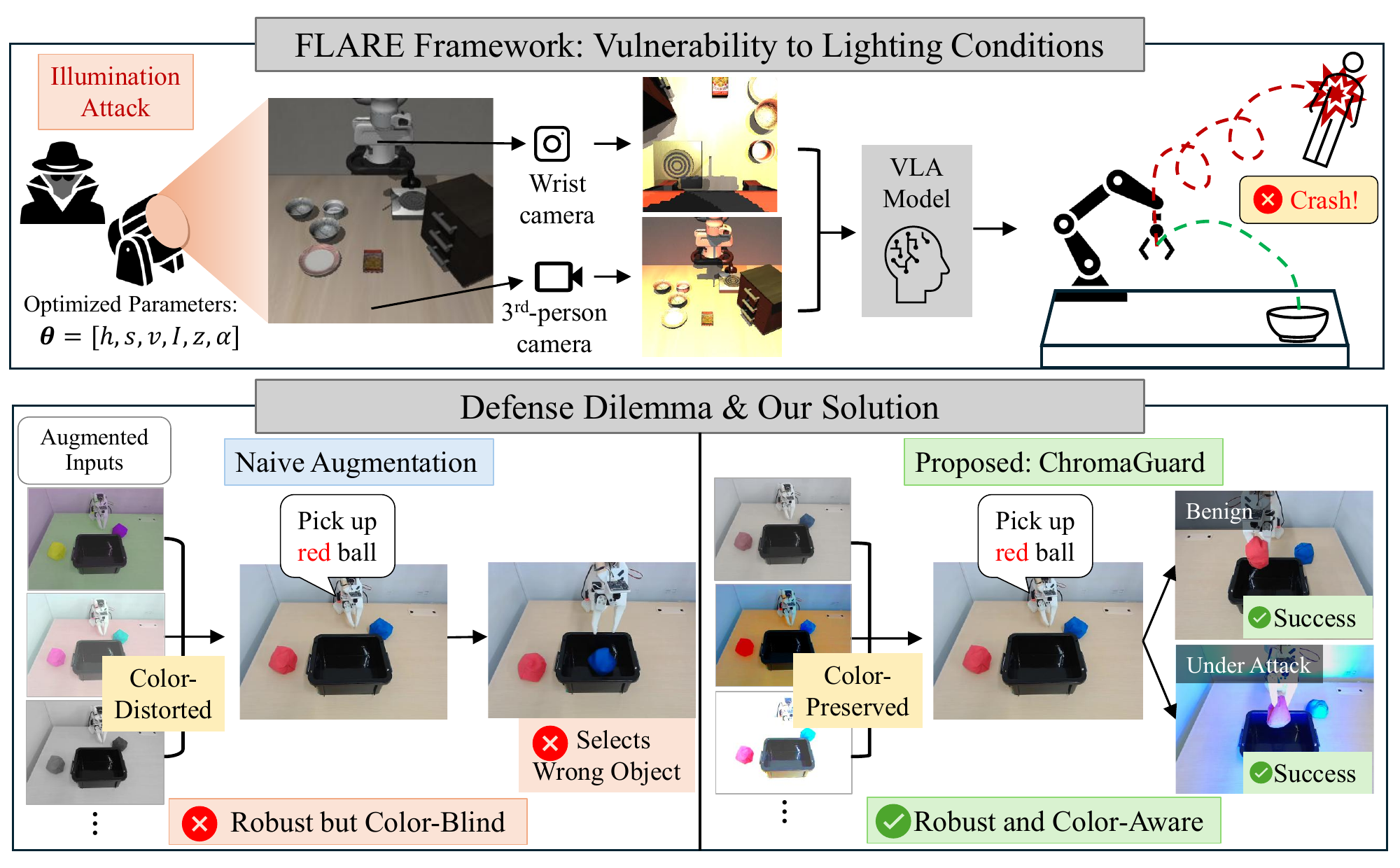}
    \caption{Full Pipeline of Our Proposed FLARE Framework and ChromaGuard Defense. Top: Our FLARE framework optimizes spotlight parameters to induce catastrophic trajectory deviations in VLA models. Bottom: The defense dilemma where naive color augmentation degrades task success on color-dependent tasks, which is resolved by our proposed ChromaGuard approach.}
    \label{fig:teaser_pipeline}
\end{figure}

Vision-language-action (VLA) models extend large-scale vision-language learning to robot control, offering a promising route toward general-purpose robot manipulation~\cite{Zitkovich2023rt-2, kim2024openvla, shukor2025smolvla, black2024pi0}. This integration empowers robotic agents to generalize across a wide variety of dynamic tasks in diverse domains, such as household manipulation~\cite{jiang2025behavior}, autonomous driving~\cite{jiang2025survey}, and industrial automation~\cite{li2025transferring}.
However, as VLAs increasingly transition from controlled laboratory environments to safety-critical, real-world deployments, their tightly coupled multimodal architectures expose them to novel security vulnerabilities~\cite{VLA_Safety_Survey, Embodied_AI_Security_Survey}. While recent research has highlighted the vulnerability of VLAs to advanced threats, such as physical sensor attacks~\cite{lu2026phantom}, adversarial attacks~\cite{EDPA_Paper, FreezeVLA}, and training-time backdoor injections~\cite{zhou2026badvla}, we find that the underlying vulnerability often stems from poor generalizability rather than security flaws requiring advanced attacker knowledge. 
To highlight this fundamental weakness, we systematically demonstrate that state-of-the-art VLA models fail on standard benchmarks when subjected to minor physical environmental variations, such as a slight shift in lighting conditions caused by a targeted physical spotlight.
Adversarial training is widely recognized as an effective countermeasure for improving the generalizability of VLA models in prior literature~\cite{lu2026phantom}. However, we find that naive adversarial training does not consistently yield genuine robustness; rather, it often conditions VLA models to largely overlook the visual features that were perturbed during the training process. For example, when applying broad HSV-space data augmentation to the training data, the resulting VLA model 
suffers a severe drop in color discrimination capabilities. Consequently, the model loses the capacity to solve semantic tasks that rely on fine-grained color cues.
To resolve this defense dilemma, we introduce ChromaGuard, a chroma-preserving adversarial training methodology, as depicted in Fig.~\ref{fig:teaser_pipeline}. We demonstrate that our approach not only provides robust defensive capabilities against severe lighting attacks but also maintains high performance in benign, color-dependent operations. Our findings highlight a critical lesson for embodied AI: standard handcrafted countermeasures can introduce biases that override fundamental semantic signals, particularly in VLAs, where specific elements of the visual observation dictate safety-critical actions. The major contributions of this study are as follows:
\begin{itemize}[itemsep=0pt, leftmargin=1.5em]
\item We systematically expose a generalization vulnerability in state-of-the-art VLAs by proposing FLARE. This optimized physical spotlight attack framework degrades task success to near 0\% in our evaluation while inducing trajectory deviations of up to 115.5 cm in simulation.
\item We identify a critical defensive trap in standard adversarial training methods, such as naive HSV transformation. Through diagnostic validation, we demonstrate that naive augmentations inadvertently condition the model to discard color cues. While the model retains up to 90.5\% success on purely grayscale inputs, this conditioning translates to a steep performance drop to no more than 48.8\% success in real-world semantic tasks requiring color differentiation.
\item We propose ChromaGuard, a novel augmentation architecture designed to preserve hue. We empirically demonstrate that this method has high defense effectiveness against adversarial lighting attacks, reaching 67.5\% success in color-invariant tasks. Furthermore, it strictly retains the perceptual capabilities necessary for complex physical manipulation by yielding success rates of 95.0\% and 88.8\% in benign and attacked color-dependent scenarios, respectively.

\end{itemize}

\section{Background}

\subsection{Generalizability and Environmental Robustness of VLA Models}
The multimodal nature of VLAs exposes a wide variety of attack channels~\cite{VLA_Safety_Survey, Embodied_AI_Security_Survey} such as training-time backdoor attacks~\cite{zhou2026badvla, INFUSE_Zhou_2026} and inference-time adversarial attacks~\cite {EDPA_Paper, ADVLA_Paper, FreezeVLA}. 
Meanwhile, these advanced attacks naturally extend the prior attacks effective against general DNN models. The major contributions of these previous works are in overcoming the major challenges in triggering the vulnerabilities in their VLA domains.  In this work, we highlight that VLA models are indeed vulnerable to these kinds of attacks, but the current state-of-the-art VLA models are poorly generalized and are just vulnerable to physical environmental variations, rather than particularly vulnerable to these advanced attacks. LIBERO-PRO~\cite{zhou2025liberopro} also reports that four major state-of-the-art VLA models easily collapse under small task and position perturbations on LIBERO~\cite{liu2023libero}, one of the most popular VLA benchmarks. All models achieve $\geq$90\% success rates on the original LIBERO tasks but fall to nearly 0\% under the perturbations, showing poor generalizability. Under these circumstances, any attacks may succeed. This motivates us to start by investigating the impact of lighting conditions, which can be perturbed in normal benign scenarios.

To handle the vulnerability of VLA systems, recent research has explored both runtime guardrails~\cite{CRT_Paper, RobustVLA_Paper} and training-time regularization~\cite{lu2026phantom}. To directly address the generalizability of DNN models, adversarial training~\cite{madry2018towards, lu2026phantom} has been widely adopted not only in VLA~\cite{lu2026phantom} but also in many other application domains~\cite{xhonneux2024efficient, ganin2016domain, hanif2023frequency}. However, we find that the naive use of adversarial training may severely compromise the model's generalizability. To be robust against lighting condition changes, a naive approach is to apply color jittering for input images, also known as one of the data augmentation techniques~\cite{yang2022image}. As our study demonstrates, this inadvertently conditions the VLA to discard essential color and texture cues, effectively degrading its visual encoder into a heavily shape-biased, grayscale processor. In complex manipulation tasks where color differentiation is critical to object interaction, this loss of semantic fidelity translates to tangible task failures. Consequently, there is an urgent need for a defensive framework that simultaneously achieves robustness against adversarial illumination while strictly preserving the integrity of critical chroma features.

\section{Methodology}

\subsection{Threat Model}\label{sec:threat_model}

We follow the common threat model adopted in prior work~\cite{lu2026phantom}, which assumes a strict black-box setup, i.e., the adversary has no access to the target model's internal architecture, weights, gradient information, or training datasets. As depicted in Fig.~\ref{fig:teaser_pipeline}, the adversary can deploy physical lighting fixtures within the robot's workspace. The adversary can control its height, illumination footprint, intensity, and the specific chromatic properties (hue, saturation, and value) of the emitted light. The planar position (x, y) of the lighting is fixed, and its orientation is directed toward the center of the workspace. However, they cannot dynamically change the light once deployed; the parameters remain fixed across all evaluation episodes.

\begin{figure*}[t]
    \centering
    \includegraphics[width=\linewidth]{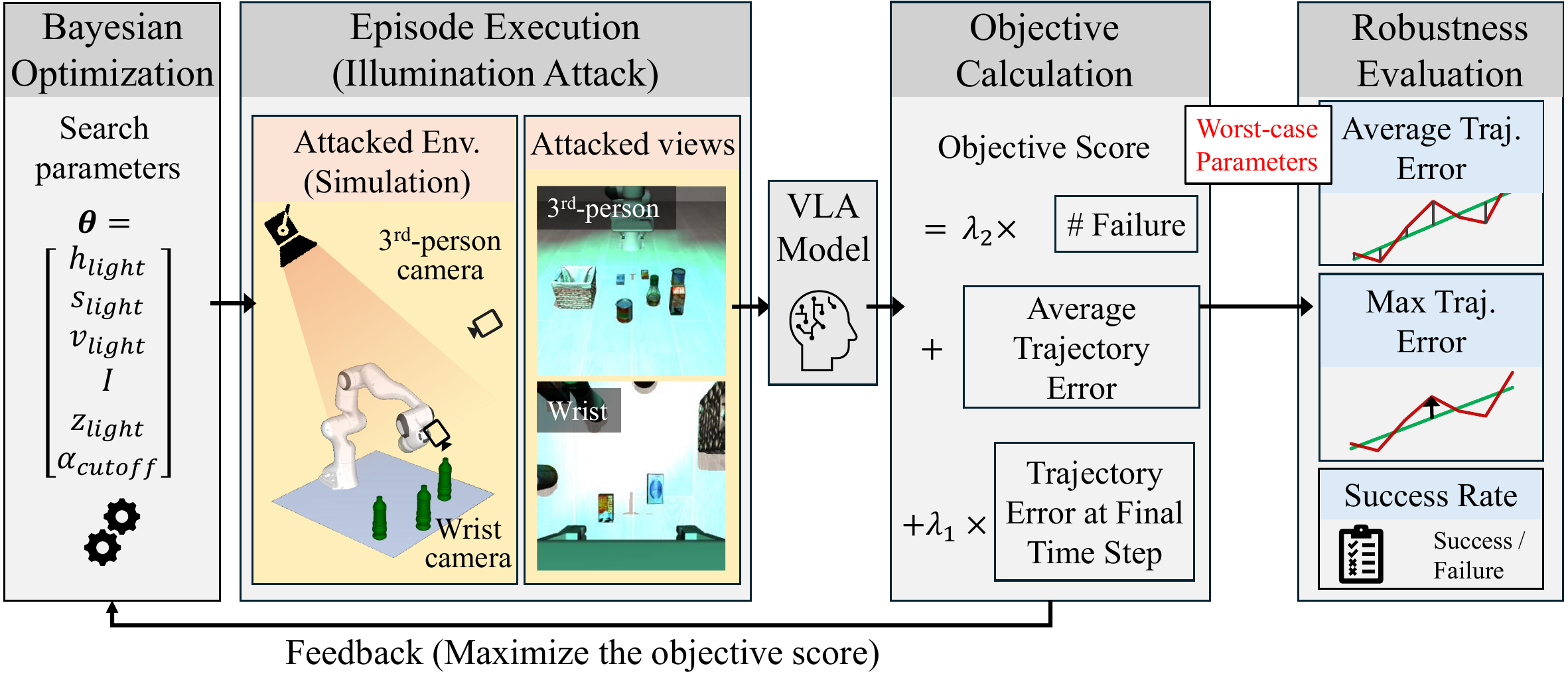}
    \caption{Overview of the FLARE (Framework for Lighting Adversarial Robustness Evaluation). The framework utilizes Bayesian Optimization to sample worst-case physical spotlight parameters, systematically evaluating VLA robustness based on trajectory deviations and task failure rates.}
    \label{fig:flare_framework}
\end{figure*}

\subsection{FLARE Framework: Optimized Physical Spotlight Attack}
We propose the Framework for Lighting Adversarial Robustness Evaluation (FLARE), illustrated in Fig.~\ref{fig:flare_framework}, to generate worst-case lighting scenarios. We consider a black-box adversary that deploys a spotlight in the workspace and configures its parameters $\theta$. The attack is physically plausible: it does not modify the robot, camera, or target objects, and operates exclusively through illumination. In simulation, the spotlight is rendered directly to optimize these parameters; on hardware, the same parameterization corresponds to a physical lamp whose height, beam angle, and color output can be adjusted within safe bounds.

We parameterize the spotlight by:
\begin{equation}
\theta = [h_{\text{light}}, s_{\text{light}}, v_{\text{light}}, I, z_{\text{light}}, \alpha_{\text{cutoff}}],
\end{equation}
where $h_{\text{light}}, s_{\text{light}}, v_{\text{light}}$ denote the hue, saturation, and value of the light source, $I$ is the light intensity, $z_{\text{light}}$ is the spatial height (z-coordinate), and $\alpha_{\text{cutoff}}$ represents the spotlight's cutoff angle.

Given a clean observation $o_t$, the attacked observation is produced by a lighting operator $o_t^{\text{adv}} = \mathcal{A}(o_t; \theta)$, which alters illumination while preserving physical scene geometry and object layout. The attack objective follows the intuition that dangerous failures are not only unsuccessful but also catastrophically distant from the intended trajectory. For episode $i$, we score a spotlight parameter by combining trajectory deviation and task failure:
\begin{equation}
L_i(\theta) = \frac{1}{S}\sum_{t=1}^{S}\lVert \mathbf{P}_t^{(i)} - \mathbf{P}_t^{\prime (i)} \rVert_2 + \lambda_1 \lVert \mathbf{P}_T^{(i)} - \mathbf{P}_T^{\prime (i)} \rVert_2 + \lambda_2 \mathbb{I}_f^{(i)},
\label{eq:objective}
\end{equation}
where $S$ is the maximum number of steps, $\mathbf{P}_t^{(i)}$ and $\mathbf{P}_t^{\prime (i)}$ are the clean and attacked end-effector positions respectively, $\mathbb{I}_f^{(i)} \in \{0,1\}$ indicates a binary task failure, and $\lambda_1, \lambda_2$ are weighting hyperparameters. We aim to maximize the average objective score over $N$ episodes: $J(\theta) = \frac{1}{N}\sum_{i=1}^{N} L_i(\theta)$.

Because physical and simulated rollouts are expensive and internal model gradients are unavailable, we employ Bayesian Optimization to efficiently search the continuous spotlight parameter space for values that maximize $J(\theta)$. Specifically, we use Optuna~\cite{akiba2019optuna} with a Tree-structured Parzen Estimator (TPE) sampler. While conventional Gaussian Process-based Bayesian Optimization (GP-BO) is widely used, its computational overhead increases significantly with the number of trials. In our context demanding hundreds of iterations of high-cost robot rollouts involving multi-modal VLA inference and physical simulation, this overhead presents a severe bottleneck. To avoid this problem, we select the TPE sampler because its computational cost is substantially lower than that of GP-BO, making it highly suitable for high-dimensional continuous search spaces under a limited time budget. 
Furthermore, evaluating all $10$ tasks within a single LIBERO task suite at every iteration is computationally impractical in terms of total execution time. To balance computational efficiency with the discovery of truly universal adversarial parameters, we evaluate the tasks in alternating subsets. At each optimization iteration, the task suite is partitioned into two distinct subsets of $5$ tasks, which are alternately evaluated to compute the trial score. This evaluation protocol prevents the optimization from overfitting to specific scenarios within a constrained computational budget. 
Detailed optimization hyperparameters and computational times are provided in Appendix~\ref{sec:optimization}.

\subsection{ChromaGuard: Hue-Preserving Adversarial Training}

Naive lighting augmentation typically perturbs all color channels indiscriminately, often through aggressive hue, saturation, or texture-distorting transforms. We first introduce a typical naive data augmentation (we call this baseline \textit{Naive-Aug}) which is a standard combination of color and sharpness jittering. We denote the augmentation parameters as $\phi = [h, s, v, c, \gamma]^\top$, representing perturbations on hue, saturation, brightness, contrast, and sharpness, respectively. These parameters are randomly sampled from uniform distributions. While this approach improves general visual robustness, unconstrained hue variations ($h \neq 0$) can alter intrinsic object colors, inadvertently confounding performance on strictly color-dependent tasks.

To isolate illumination effects, \textit{ChromaGuard} restricts the augmentation parameter space. Specifically, it allows bounded perturbations to saturation, brightness, contrast, and sharpness, but explicitly fixes the hue perturbation to zero ($h = 0$). We use this constrained parameter space, denoted as $\Phi_{\text{CG}}$. To simulate challenging lighting conditions, we formulate an adversarial objective: $\min_{w} \mathbb{E}_{x, y} \left[ \max_{\phi \in \Phi_{\text{CG}}} \mathcal{L}(f_{w}(T(x; \phi)), y) \right]$. Since solving this exact inner maximization is computationally prohibitive for VLAs, we practically approximate it via uniform random sampling, relying on our broad bounds to capture worst-case illuminations.

\section{Simulator Experiments: Exposing Vulnerability against Spotlight Attack} \label{sec:experiments}

We evaluate the performance of our VLA model against spotlight attacks within a simulated environment. In this section, we demonstrate that (1) even state-of-the-art VLA models can be easily compromised by simple spotlight illumination, and (2) a high degree of robustness against this attack can be seemingly achieved through naive data augmentation, to motivate the real-world experiments (\S\ref{sec:real_world_exp}), which reveal that naive augmentation merely causes the model to ignore color information.

\subsection{Experimental Setup} \label{sec:experimental_setup_sim}

We conduct experiments within the LeRobot framework~\cite{cadene2024lerobot}, utilizing the widely adopted LIBERO multi-task manipulation benchmark~\cite{liu2023libero}. 
We train the SmolVLA architecture~\cite{shukor2025smolvla} for 200,000 steps. While the vision and language backbones utilize standard pre-trained weights, the action expert is trained from scratch. Detailed information regarding the computing resources and total training hours is provided in Appendix~\ref{sec:computing_resources}.
In this configuration, we train a \textit{Baseline} model with no data augmentation and a \textit{Naive-Aug} model with data augmentation that perturbs input images across multiple visual properties: specifically, $\pm 180^\circ$ in hue, $[0.0, 4.0]$ in saturation, $[0.2, 3.0]$ in value, $[0.8, 1.2]$ in contrast, and $[0.5, 1.5]$ in sharpness. Detailed model architectures, training configurations, and the design choices for these augmentation parameter ranges are provided in Appendix \ref{sec:model_architecture}.

We then evaluate these models across three diverse task suites: LIBERO-Spatial, LIBERO-Object, and LIBERO-10, conducting 500 evaluation episodes for each suite. To examine whether this vulnerability generalizes beyond SmolVLA, we additionally evaluate the publicly released OpenVLA-7B~\cite{kim2024openvla} on each LIBERO suite. The spotlight attack is optimized in a continuous parameter space spanning spatial height ($z_{\text{light}} \in [0.5, 3.0]$), full HSV color bounds, illumination intensity, and cutoff angles, using the MuJoCo physics engine~\cite{todorov2012mujoco}. 
We enabled shadow and reflection rendering within the MuJoCo engine underlying the robosuite environment to simulate realistic lighting effects. While these features are disabled by default in the standard LIBERO configurations and remained disabled during model training, we activated them across all evaluation scenarios, both benign and attacked.
For the random attack, parameters are sampled uniformly at random for each of the 500 evaluation episodes per suite.

\noindent\textbf{Evaluation Metric.} We quantify model robustness using two metrics. The \textit{Task Success Rate (SR)} measures the proportion of successfully completed episodes, serving as the baseline for operational reliability. To capture the severity of adversarial deviations, we measure the \textit{Trajectory Error (TE)} defined as the Euclidean distance between the nominal execution path (benign) and the perturbed path under attack. We report both the Average Trajectory Error (TE-Avg) across the entire episode and the Maximum Trajectory Error (TE-Max), which serves as a critical safety proxy for worst-case physical collisions. To ensure that the trajectory metrics reflect pure adversarial influence rather than inherent model incompetence, TE is strictly computed over episodes where the baseline model succeeds but fails under attack.

\subsection{Results and Implications} 

Table~\ref{tab:attack_results} shows the results of the spotlight attack against the baseline and the naively data-augmented model. As shown, the spotlight attack significantly degrades the SR of the baseline model to effectively zero. Even random spot lighting can drop the SR to less than half of the standard laboratory lighting (Benign) scenarios, where the SmolVLA-Baseline model achieves a high 81.2\%/89.4\%/58.4\% success rates for each task suite. Beyond mere task failure, the attack induces severe physical hazards: the maximum trajectory error (TE-Max) reaches $\leq$115.5 cm, forcing the robotic arm far outside its intended operational envelope. To confirm that this vulnerability is a general limitation rather than specific to SmolVLA, we conduct additional evaluations on the OpenVLA architecture, which exhibits similar catastrophic failures (see Appendix~\ref{sec:OpenVLA_result}). We also conduct an ablation study of our objective function design. Details are in Appendix~\ref{sec:objective_ablation}.

Based on the results, the naive data augmentation looks seemingly sufficient to defend against the spotlight attack. In the next section, we will demonstrate that this numerical resilience is highly deceptive, i.e., the VLA effectively degrades into a shape-biased, grayscale processor. By indiscriminately perturbing color channels during training, the augmentation forces the model to bypass the illumination challenge by discarding essential color cues. 

\begin{table*}[t]
\caption{Quantitative evaluation of the FLARE spotlight attack across three LIBERO simulation suites. The optimized attack significantly compromises the Baseline model while inducing severe trajectory deviations of up to 115.5 cm.}
\small
\label{tab:attack_results}
\begin{center}
\setlength{\tabcolsep}{4.5pt}
\begin{tabular}{lcccccc}
\toprule
\textbf{Task Suite} & \textbf{Model} & \textbf{Attack} & \textbf{SR (\%)} & \textbf{TE-Avg (cm)} & \textbf{TE-Max$_{\text{avg}}$ (cm)} & \textbf{TE-Max (cm)} \\
\hline
\multirow{6}{*}{LIBERO-Spatial} & \multirow{3}{*}{Baseline} & Benign & 81.2 & \cellcolor{lightgray}-- & \cellcolor{lightgray}-- & \cellcolor{lightgray}-- \\
& & Random & 15.4 & 12.0 & 21.7 & 37.6 \\
& & Optimized & 0.0 & 22.3 & 50.5 & 67.6 \\
\cline{2-7}
& \multirow{3}{*}{Naive-Aug} & Benign & 79.4 & \cellcolor{lightgray}-- & \cellcolor{lightgray}-- & \cellcolor{lightgray}-- \\
& & Random & 77.4 & 4.3 & 16.9 & 33.1 \\
& & Optimized & 78.8 & 4.2 & 17.8 & 40.3 \\
\hline
\multirow{6}{*}{LIBERO-Object} & \multirow{3}{*}{Baseline} & Benign & 89.4 & \cellcolor{lightgray}-- & \cellcolor{lightgray}-- & \cellcolor{lightgray}-- \\
& & Random & 7.8 & 18.3 & 44.6 & 54.7 \\
& & Optimized & 0.0 & 29.1 & 75.2 & 88.3 \\
\cline{2-7}
& \multirow{3}{*}{Naive-Aug} & Benign & 89.8 & \cellcolor{lightgray}-- & \cellcolor{lightgray}-- & \cellcolor{lightgray}-- \\
& & Random & 91.6 & 7.9 & 29.3 & 56.0 \\
& & Optimized & 93.2 & 5.0 & 19.6 & 41.8 \\
\hline
\multirow{6}{*}{LIBERO-10} & \multirow{3}{*}{Baseline} & Benign & 58.4 & \cellcolor{lightgray}-- & \cellcolor{lightgray}-- & \cellcolor{lightgray}-- \\
& & Random & 22.4 & 18.5 & 37.0 & 61.6 \\
& & Optimized & 0.0 & 36.3 & 83.8 & 115.5 \\
\cline{2-7}
& \multirow{3}{*}{Naive-Aug} & Benign & 50.2 & \cellcolor{lightgray}-- & \cellcolor{lightgray}-- & \cellcolor{lightgray}-- \\
& & Random & 49.2 & 9.0 & 39.0 & 61.5 \\
& & Optimized & 47.2 & 9.0 & 40.3 & 64.0 \\
\toprule
\end{tabular}
\end{center}
\end{table*}

\section{Real-World Experiments}\label{sec:real_world_exp}

To validate the practical implications of our simulated findings, we deploy the VLA models onto a physical robotic platform. In this section, we first provide diagnostic evidence of the semantic degradation caused by naive data augmentation. We then evaluate the models across both color-invariant and color-dependent physical manipulation tasks to demonstrate that our ChromaGuard can effectively address the limitation.

\subsection{Experimental Setup} \label{sec:experimental_setup_real}

\begin{figure}[t]
    \centering
    \includegraphics[width=\linewidth]{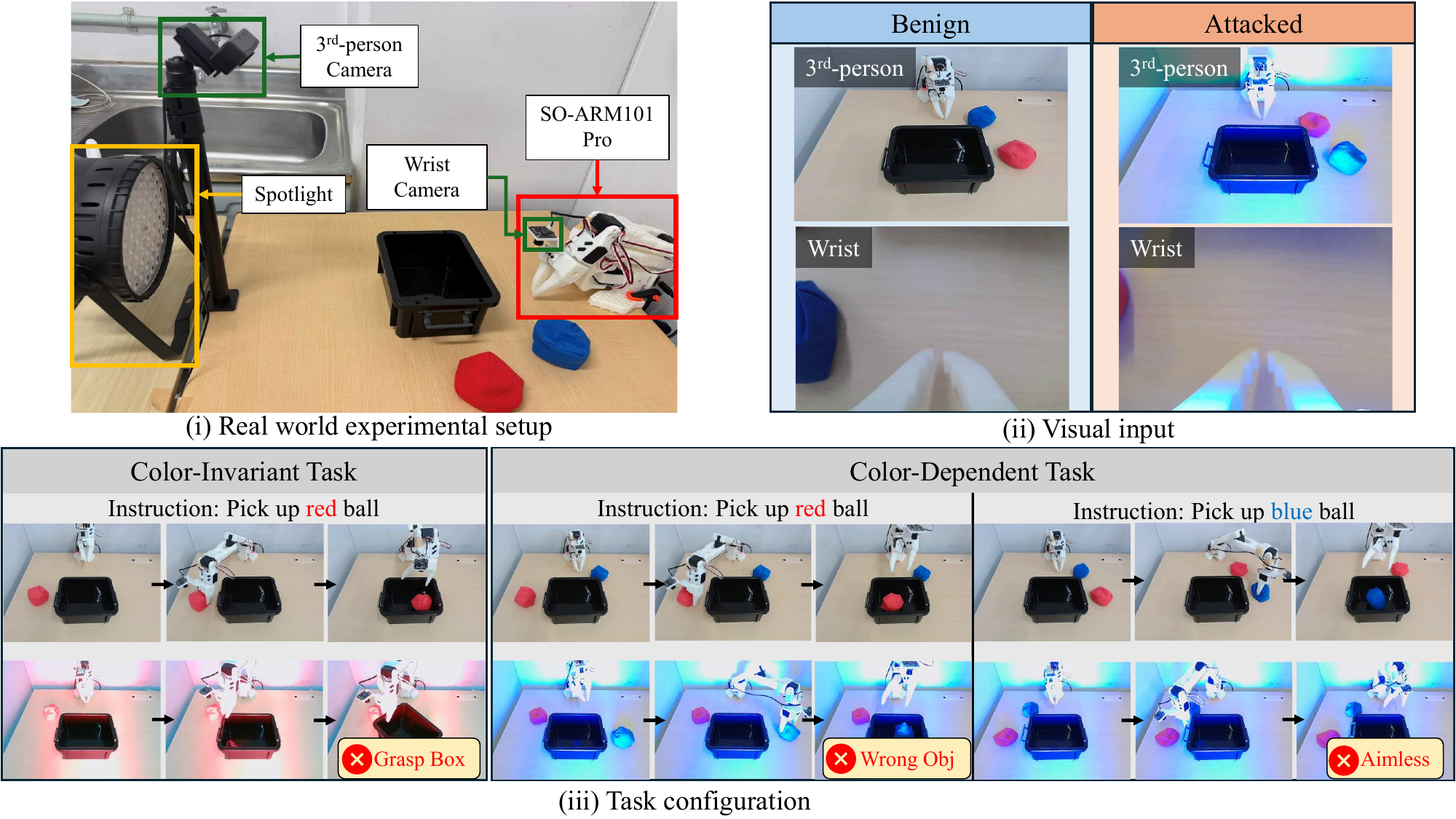}
    \caption{Real-world Experimental Setup. (i) The physical workspace featuring a 6-DoF robotic arm, dual-camera system, and adjustable spotlight. (ii) Examples of visual inputs from 3rd-person and wrist cameras under benign and attacked illumination. (iii) Task configurations comparing color-invariant and color-dependent manipulation instructions.}
    \label{fig:setup}
\end{figure}

Fig.~\ref{fig:setup} illustrates our experimental setup, which employs a 6-DoF SO-101 Arm Pro manipulator~\cite{Knight_Standard_Open_SO-100} equipped with a dual-camera perception system, comprising a wrist-mounted camera and a static third-person camera. 
To accurately reflect practical camera adaptations, both auto-exposure and auto-white-balance were enabled throughout all physical experiments.
To synthesize the spotlight attack, an external, programmable RGB spotlight is positioned outside the camera's field of view. 
Naturally, shadows and reflections are considered in the physical real-world experiments.
We define two task configurations: a color-invariant task (e.g., picking a red ball and placing it into a black bin) and a highly color-dependent task (e.g., selecting between structurally identical red and blue balls based on language commands). Further details on the task setups and data collection protocols are provided in Appendix~\ref{sec:data_collection}.

To perform these tasks, we train both the SmolVLA~\cite{shukor2025smolvla} and $\pi_{0.5}$ ~\cite{black2025pi_} architectures by fine-tuning the \texttt{lerobot/smolvla\_base} and \texttt{lerobot/pi05\_base} checkpoints, respectively. Detailed architectural differences and input configurations between these two models are provided in Appendix~\ref{sec:model_architecture}.

Both models are fine-tuned for 100,000 steps using our collected real-world demonstrations, comprising 50 episodes for the color-invariant task and 240 episodes for the color-dependent task. As our adversarial training approach, \textit{ChromaGuard} fixes hue and perturbs $[0.0, 4.0]$ in saturation, $[0.2, 3.0]$ in value, $[0.8, 1.2]$ in contrast, and $[0.5, 1.5]$ in sharpness during the model training. We use the same parameters for the \textit{Baseline} and \textit{Naive-Aug} as those used in~\S\ref{sec:experimental_setup_sim}.
For the spotlight attack generation, we first generate several attack candidates in the simulator with our FLARE optimization while limiting our available colors (red, blue, and green), and then we select the most effective attack in the real-world.
We evaluate the trained models by conducting 40 trials for the color-invariant task. For the color-dependent task, we conduct a total of 80 trials, consisting of 40 trials for each of the two language instructions.

\subsection{Evaluation of Color Reliance via Grayscale Inference}

To systematically verify whether the augmented model loses its reliance on color information, we first conduct a grayscale diagnostic evaluation. We convert the input images to grayscale during inference across three different LIBERO task suites. As shown in Table~\ref{tab:grayscale_eval}, the baseline model, which naturally learned color features during training, suffers a significant drop, its success rates falling to near zero across the suites. In contrast, the Naive-Aug model maintains almost identical performance on grayscale inputs as on standard RGB, e.g., 90.5\% vs. 89.8\% SR on LIBERO-Object. 

This phenomenon exposes a critical flaw in naive data augmentation. Because tasks rarely require disambiguating identically shaped objects solely by color in standard benchmarks, the Naive-Aug model learns to discard color channels entirely as noisy variables. It bypasses the illumination challenge not by learning robust color representations, but by collapsing its perception into a shape- and position-biased state, effectively ignoring semantic color cues.

\begin{table}[t]
\centering
\caption{Grayscale diagnostic evaluation. The high success rate of Naive-Aug on artificially desaturated images shows an over-reliance on shape biases, indicating semantic degradation of color cues.}
\label{tab:grayscale_eval}
\begin{tabular}{llcc}
\toprule 
\multirow{2}{*}{\textbf{Task Suite}} & \multirow{2}{*}{\textbf{Model}} & \multicolumn{2}{c}{\textbf{Task Success Rate (\%)}} \\
\cline{3-4}
& & \textbf{Benign (RGB)} & \textbf{Grayscale} \\
\hline
\multirow{2}{*}{LIBERO-Spatial} & Baseline & 81.2 & 0.0 \\
& Naive-Aug & 79.4 & 80.1 \\
\hline
\multirow{2}{*}{LIBERO-Object} & Baseline & 89.4 & 10.0 \\
& Naive-Aug & 89.8 & 90.5 \\
\hline
\multirow{2}{*}{LIBERO-10} & Baseline & 58.4 & 0.0 \\
& Naive-Aug & 50.2 & 47.8 \\
\toprule 
\end{tabular}
\end{table}

\begin{table}[t]
\centering
\caption{Real-world evaluation of model robustness. Unlike naive augmentation, ChromaGuard maintains high perceptual fidelity in benign scenarios while improving robustness against targeted physical spotlight attacks. Numbers in parentheses indicate the percentage of total failures caused by grasping incorrectly colored objects. The $\pm$ values denote the larger of the lower and upper margins of the 95\% Wilson score interval.}
\label{tab:real_world_eval}
\small
\setlength{\tabcolsep}{9pt}
\begin{tabular}{llcccc}
\toprule
\multirow{2}{*}{\textbf{Base Model}} & \multirow{2}{*}{\textbf{Model}} 
  & \multicolumn{2}{c}{\textbf{Color-Invariant Task SR}} 
  & \multicolumn{2}{c}{\textbf{Color-Dependent Task SR}} \\
\cline{3-6}
  & & \textbf{Benign} & \textbf{Attack} & \textbf{Benign} & \textbf{Attack} \\
\hline
\multirow{3}{*}{SmolVLA}
  & Baseline    & 67.5±15.5 & 0.0+8.8 & 78.8±10.2 (11.8)& 21.3±10.2 (22.2) \\
  & Naive-Aug           & 77.5±15.0 & \textbf{67.5±15.5} & 48.8±10.8 (90.2) & 37.5±11.0 (80.0) \\
  & \textbf{Ours}  & \textbf{85.0±14.1} & \textbf{67.5±15.5} & \textbf{95.0±7.2} (0) & \textbf{88.8±8.8} (22.2)\\
\hline
\multirow{3}{*}{\textbf{$\pi_{0.5}$}}
  & Baseline    & 77.5±15.0 & 25.0±15.2 & 47.5±10.8 (83.3) & 18.8±9.9 (35.4)\\
  & Naive-Aug           & 77.5±15.0 & \textbf{77.5±15.0} & 40.0±11.0 (79.2)& 47.5±10.8 (88.1)\\
  & \textbf{Ours}  & \textbf{80.0±14.8} & \textbf{77.5±15.0} & \textbf{51.3±10.8} (100)& \textbf{55.0±10.9} (91.7)\\
\toprule
\end{tabular}
\end{table}
 
\subsection{Evaluation of Color-Dependent and Color-Invariant Tasks}
\label{sec:color_tasks}
To assess the practical impact of augmentation-induced perceptual degradation, we evaluated the models on physical hardware across color-invariant and color-dependent tasks. Table~\ref{tab:real_world_eval} shows the success rates of the color-invariant and color-dependent tasks for each model. As shown,  in the color-invariant scenario, our ChromaGuard and Naive-Aug models defended against the spotlight attack as expected based on the results in~\S\ref{sec:experiments}. The success rates of the SmolVLA and $\pi_{0.5}$ improve from 0\% to 67.5\% and from 25.0\% to 77.5\% under attack, respectively. 

However, in the color-dependent scenario, where resolving the semantic instruction relies entirely on visual color cues, the Naive-Aug model generalized poorly. Even under benign laboratory lighting, its success rate dropped to 48.8\% compared to the Baseline's 78.8\%, as the robot frequently struggled to distinguish between the identical red and blue objects. This confirms that defense techniques relying on indiscriminate color perturbation can severely degrade the semantic grounding of color information required for complex real-world tasks.

ChromaGuard resolves this trade-off by decoupling illumination variations from chroma features during training. As shown in Table~\ref{tab:real_world_eval}, ChromaGuard matches the robustness of Naive-Aug in the color-invariant setting ($\geq$67.5\% SR under attack) while demonstrating substantial improvements in the color-dependent setting. Specifically, it successfully preserves semantic fidelity: SmolVLA achieves a 95.0\% SR under benign conditions and 88.8\% under attack, while $\pi_{0.5}$ yields 51.3\% and 55.0\% SR, respectively.
We hypothesize that the limited color discrimination of $\pi_{0.5}$, even under benign conditions, stems from the pre-training of its foundation model rather than the augmentation strategy. As evidence, $\pi_{0.5}$ frequently grasps objects of the incorrect color even in benign scenarios. This factor accounts for 83.3\% of all failure cases, a rate seven times higher than the corresponding failure rate (11.8\%) observed in SmolVLA; furthermore, ChromaGuard achieves a benign success rate of 51.3\%, on par with the baseline (47.5\%), and maintains its performance under attack (55.0\%).

\noindent\textbf{Ablation Study.} We further validate ChromaGuard on SmolVLA with two ablations (Appendix~\ref{sec:ablation}). On a harder color-dependent task that requires distinguishing red from pink objects with similar hues, ChromaGuard achieves the highest SR in both benign and attacked scenarios (82.5\% and 72.5\%), whereas Naive-Aug drops to 50.0\% even in benign conditions. In addition, fixing any additional property alongside hue or narrowing the perturbation range does not improve robustness, while none of the variants grasps an incorrectly colored object in benign scenarios.

\section{Limitations and Discussions}

Although our FLARE and ChromaGuard demonstrate substantial attack and mitigation capabilities on spotlight attacks, we do not intend them to serve as definitive attack and defense methods. Instead, they serve as analytical tools to expose the acute vulnerability of VLA models to subtle environmental perturbations. Our analysis reveals that naive data augmentation fails to resolve these vulnerabilities, merely discarding color features rather than learning robust representations. Furthermore, these tools do not yet explore temporal dynamics, such as time-varying lighting or closed-loop adaptive attacks. Our primary objective is to expose the critical robustness crisis facing current VLA models, rather than to engineer an exhaustive adversarial attack or an impenetrable defense.

While our evaluation focuses on static spotlight attacks, real-world robotic perception is also known to be vulnerable to other perturbations. Specifically, although our attack induces complex variations, we have not isolated and evaluated shadows, reflections, or motion blur as independent perturbation factors. Recent studies have shown that randomized smoothing can provide certified robustness guarantees against complex perturbations, such as camera motion~\cite{hu2022robustness, hu2024pixel}. Since ChromaGuard operates at the pixel level during training, it could potentially be combined with such certified defenses, which we leave for future work.

\section{Conclusion}

We investigated the robustness of Vision-Language-Action (VLA) models under targeted physical illumination variations. Utilizing our proposed FLARE evaluation framework, we demonstrated that optimized spotlight perturbations can degrade baseline VLA performance to a near-zero success rate while inducing substantial trajectory deviations. Furthermore, we identified a practical limitation in standard color data augmentations: while they successfully improve robustness to lighting changes, they inadvertently bias the model to discard semantic color cues, leading to performance degradation in physical tasks that rely on color discrimination. To address this limitation, we introduced ChromaGuard, a hue-preserving augmentation strategy. Our physical evaluations on a 6-DoF manipulator indicate that ChromaGuard effectively balances robustness and perceptual fidelity, maintaining reliable success rates under targeted lighting attacks without compromising the visual acuity necessary for semantic manipulation. These findings emphasize the urgent necessity of achieving a certain level of generalizability of the current state-of-the-art VLA models before diving too deeply into their security against sophisticated attacks.

\acknowledgments{We thank our anonymous reviewers for their insightful feedback and Taiki Miyanishi for discussions on the VLA models. This research was supported in part by JST Next-generation Edge AI Semiconductor
JPMJES2515, JST CREST JPMJCR23M4, JST PRESTO JPMJPR22PA, JST ASPIRE JPMJAP2515, JSPS KAKENHI 24K02940.}

\clearpage
\bibliography{example}

\appendix
\clearpage
\section*{Appendix}

\renewcommand{\thesection}{\Alph{section}}

\title{Supplementary Material:\\ Lights, Camera, Malfunction: When Illumination Robustness Leaves VLA Models Blind to Color}

\author{}

\maketitle

\section{Optimization Details of the Illumination Attack}\label{sec:optimization}

For the Bayesian Optimization process utilizing the Tree-structured Parzen Estimator (TPE) sampler introduced in~\S3.2, we set the total number of optimization trials to 300. To ensure sufficient initial random exploration of the parameter space, the number of startup trials is configured to 50. We set $\lambda_1 = 2.0, \lambda_2 = 5.0$.

The computational time required for the simulation-based optimization and evaluation phases, conducted on a single NVIDIA RTX A5000 GPU, varies across the LIBERO task suites. This variation is primarily driven by the maximum environment steps allowed per episode, which are defined as 220, 280, and 520 steps for the LIBERO-Spatial, LIBERO-Object, and LIBERO-10, respectively. Consequently, the optimization process required approximately 4--8 hours for Spatial, 4--11 hours for Object, and 13--14 hours for LIBERO-10. The subsequent evaluation phase took approximately 2.5--3.5 hours, 2--6.5 hours, and 8--11.5 hours for the respective suites. For the real-world manipulator experiments, each physical rollout is executed with a time limit of up to 15 seconds.

\section{Computing Resources for Training}\label{sec:computing_resources}

All models were trained using a single NVIDIA RTX A5000 or RTX Pro 6000 GPU. For the simulation experiments, training the SmolVLA architecture for 200,000 steps consistently required approximately 114 hours (about 4.7 days) per model on an RTX A5000. This training duration remained stable regardless of the task suite~\cite{liu2023libero} (LIBERO-Spatial, LIBERO-Object, and LIBERO-10) or the presence of data augmentation.
For the real-world manipulator experiments, training the SmolVLA models for 100,000 steps on an RTX A5000 took approximately 2 days per model. Conversely, training the $\pi_{0.5}$ models on an RTX Pro 6000 required approximately 42 hours (about 1 day and 18 hours). In all cases, ChromaGuard did not introduce any significant computational overhead, which is at the same level as the normal data augmentation methods.

\section{Details of Model Architecture and Training Configuration}\label{sec:model_architecture}

Our study employs two major VLA architectures: SmolVLA~\cite{shukor2025smolvla} and  $\pi_{0.5}$~\cite{black2025pi_}. For SmolVLA, we use the same architectural configurations and training hyperparameters for both simulation and real-world experiments. 
All model architectures and training pipelines are implemented utilizing the open-source LeRobot~\cite{cadene2024lerobot} framework.
For the SmolVLA configuration, we utilize the pre-trained \texttt{HuggingFaceTB/SmolVLM2-500M-Video-Instruct} as the vision-language backbone. Given multi-view visual observations, the robot's internal state, and a language instruction, the model fuses these inputs to predict a chunk of continuous robot actions.
At inference time, the policy predicts action chunks of length 50.  We freeze the vision encoder and train the state projection layer and the action expert from scratch.  

Similarly, the $\pi_{0.5}$ architecture follows the exact same input-output formulation, taking multi-view images, robot's internal states, and language instructions to generate action chunks of length 50.
Its backbone is configured with the PaliGemma \texttt{gemma\_2b} variant.
In contrast to SmolVLA, $\pi_{0.5}$ leverages a pre-trained language model, specifically the \texttt{gemma\_300m} architecture as the action expert. 
We train SmolVLA and  $\pi_{0.5}$ by using the AdamW optimizer with a cosine learning rate decay scheduler following a 1,000-step linear warmup phase. To accommodate the input requirements of each vision encoder, the raw visual observations are padded and resized to $512 \times 512$ pixels for the SmolVLA and resized to $224 \times 224$ pixels for the $\pi_{0.5}$. 

The augmentation parameter ranges for Naive-Aug were defined by expanding the default data augmentation settings of the LeRobot framework to their maximum viable limits. Crucially, these limits were carefully calibrated to preserve visual perceptibility, preventing severe image degradation such as complete overexposure (whiteout) or underexposure (blackout).
Specifically, these transformations consist of both color and sharpness jittering. 
The ranges for hue, saturation, and brightness were significantly expanded from their defaults of $\pm 18^\circ$, $[0.5, 1.5]$, and $[0.8, 1.2]$ to the values detailed in the main paper. Meanwhile, the parameters for contrast and sharpness were maintained at their default ranges of $[0.8, 1.2]$ and $[0.5, 1.5]$. In Appendix~\ref{sec:ablation}, these original default bounds are utilized as the parameters for the narrow-range variant.

\section{Vulnerability of Additional VLA Architectures}\label{sec:OpenVLA_result}
To demonstrate that the vulnerability to physical illumination changes is not specific to the model architectures evaluated in the main text, we conducted additional simulation evaluations using OpenVLA~\cite{kim2024openvla}, specifically the publicly available 7B parameter model (\texttt{openvla-7b-finetuned-libero-spatial/object/10}). We evaluated the model's robustness against the optimized spotlight setting generated by FLARE across three LIBERO task suites~\cite{liu2023libero}: LIBERO-Spatial, LIBERO-Object, and LIBERO-10.
As shown in Table~\ref{tab:openvla_result}, OpenVLA exhibits severe performance degradation under the targeted lighting variations, demonstrating similar fragility to SmolVLA. Specifically, the task success rate drops from 84.0\% to as low as 1.6\% on LIBERO-Spatial. Furthermore, the model experiences significant trajectory deviations, with maximum average trajectory errors reaching up to 42.3 cm on the LIBERO-Object suite.
These results confirm that the vulnerability against physical illumination changes is a fundamental and general limitation across state-of-the-art VLA architectures.

\begin{table}[htb]
    \centering
    \caption{OpenVLA performance degradation under FLARE perturbations on LIBERO suites.}
    \label{tab:openvla_result}
    \begin{tabular}{lccccc}
    \toprule
    \textbf{Task} & \textbf{Attack} &\textbf{SR (\%)} & \textbf{TE-Avg (cm)} 
    & \textbf{TE-Max$_{\text{avg}}$ (cm)} & \textbf{TE-Max (cm)} \\
    \hline
    \multirow{2}{*}{LIBERO-Spatial} & Benign & 84.0 & \cellcolor{lightgray}-- & \cellcolor{lightgray}-- &  \cellcolor{lightgray}-- \\
    & Optimized & 1.6  & 15.1 & 27.1 & 51.2 \\
    \hline
    \multirow{2}{*}{LIBERO-Object} & Benign & 64.8 & \cellcolor{lightgray}-- & \cellcolor{lightgray}-- &  \cellcolor{lightgray}-- \\
    & Optimized & 11.2 & 18.3 & 42.3 & 51.1 \\
    \hline
    \multirow{2}{*}{LIBERO-10} & Benign & 54.8 & \cellcolor{lightgray}-- & \cellcolor{lightgray}-- &  \cellcolor{lightgray}-- \\
    & Optimized & 10.4 & 19.8 & 36.0 & 43.8 \\
    \toprule
    \end{tabular}%
\end{table}

\section{Ablation Study of the Objective Function Design} \label{sec:objective_ablation}

To justify the specific mathematical formulation of our attack, we conducted an ablation study isolating the components of our proposed objective function (Eq. 2). As demonstrated in Table~\ref{tab:objective_ablation}, every ablated formulation successfully forces a complete task failure, reducing the Success Rate (SR) to $0.0\%$. However, achieving task failure alone does not maximize the physical hazard.
Removing the Average Trajectory Error reduces the overall deviation of the arm, yielding the lowest values across all evaluation metrics.
Similarly, excluding the Final Position Error or the Task Failure Flag consistently diminishes the absolute maximum trajectory error. These results confirm that forcing a semantic task failure is deeply correlated with maximizing physical displacement; the complete tripartite objective strikes an optimal balance, ensuring the attack is both highly successful at halting the task and maximally disruptive to physical safety bounds.

\begin{table*}[h]
    \centering 
    \caption{Ablation study of the adversarial objective function (Eq. 2). Removing individual terms visibly limits the attack's ability to maximize trajectory deviations compared to our full tripartite objective.} 
    \label{tab:objective_ablation} 
    \begin{tabular}{lcccc} 
    \toprule 
    \textbf{Excluded Term} & \textbf{SR (\%)} & \textbf{TE-Avg (cm)} & \textbf{TE-Max$_{\text{avg}}$ (cm)} & \textbf{TE-Max (cm)} \\ 
    \hline
    No Average Trajectory Error & 0.0 & 27.3 & 72.3 & 87.5\\
    No Final Position Error & 0.0 & 27.4 & 72.3 & 87.6 \\
    No Task Failure Flag & 0.0 & 27.8 & 73.7 & 88.0 \\
    \hline
    \textbf{Full Objective (Ours)} & \textbf{0.0} & \textbf{29.1} & \textbf{75.2} & \textbf{88.3} \\
    \toprule 
    \end{tabular} 
\end{table*}

\begin{table}[t]
\centering
    \begin{minipage}[t]{0.48\linewidth}
      \centering
        \caption{Success rates on the real-world similar-hue task (red vs. pink). ChromaGuard successfully enhances color discrimination. Parentheses as in Table~\ref{tab:real_world_eval}. Results over 40 trials.}
        \label{tab:similar_hue}
        \begin{tabular}{lcc}
        \toprule 
        \multirow{2}{*}{\textbf{Setting}} & \multicolumn{2}{c}{\textbf{Task Success Rate (\%)}} \\
        \cline{2-3}
         & \textbf{Benign} & \textbf{Attacked} \\
        \hline
        Baseline       & 65.0 (92.9) & 20.0 (37.5)\\
        Naive-Aug     & 50.0 (75.0) & 37.5 (76.0)\\
        \textbf{Ours} & \textbf{82.5} (42.9)& \textbf{72.5} (100)\\
        \toprule 
        \end{tabular}
      \includegraphics[width=0.48\linewidth]{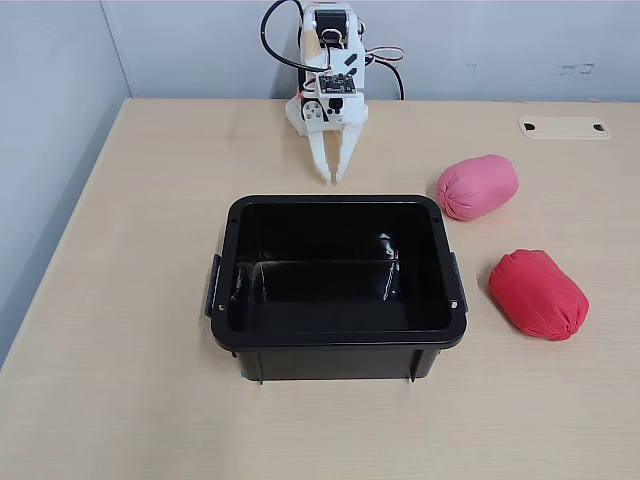}\hfill
      \includegraphics[width=0.48\linewidth]{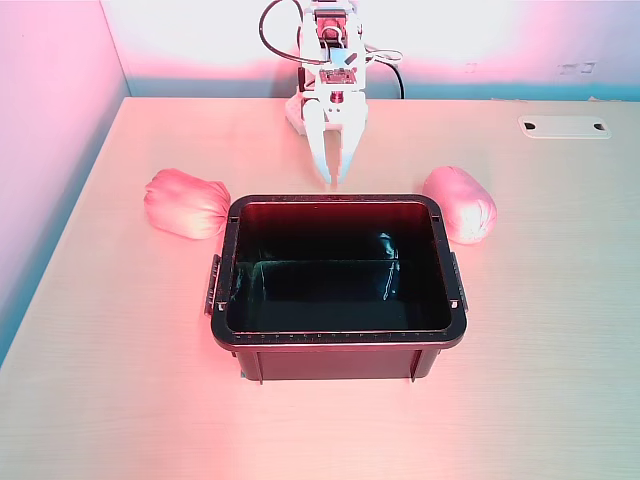}
      \captionof{figure}{Visual inputs for similar hue task under benign (left) and attacked (right) conditions.}
      \label{fig:similar_hue}
    \end{minipage}
    \hfill
    \begin{minipage}[t]{0.48\linewidth}
        \caption{Ablation of ChromaGuard on color-dependent task (SmolVLA). Ours achieves the best balance: 95.0\% benign performance and the highest attacked robustness (88.8\%). Fixing any additional visual property alongside hue, or utilizing a narrow perturbation range, does not improve robustness. Note that the ablation variants are evaluated over 40 trials each, whereas ours is evaluated over 80 trials.}
        \label{tab:ablation_aug}
        \begin{tabular}{lcc}
        \toprule
        \multirow{2}{*}{\textbf{Setting}}
        & \multicolumn{2}{c}{\textbf{Task Success Rate (\%)}} \\
        \cline{2-3}
        & \textbf{Benign} & \textbf{Attacked} \\
        \hline
        Fix H+V      & 67.5 (0) & 60.0 (37.5) \\
        Fix H+Sharp.  & 97.5 (0) & 62.5 (20.0) \\
        Fix H+Cont.   & 90.0 (0) & 75.0 (20.0) \\
        Fix H+Sat. & 92.5 (0) & 87.5 (20.0) \\
        Narrow range   & 82.5 (0) & 45.0 (36.4) \\
        \hline
        \textbf{Ours}  & 95.0 (0) & 88.8 (22.2) \\
        \bottomrule
        \end{tabular}%
    \end{minipage}
\end{table}

\section{Details of Real-World Data Collection Protocol}\label{sec:data_collection}

As briefly outlined in~\S5.1, the dataset evaluated in Section~5 was collected using a physical 6-DoF SO-101 Arm Pro~\cite{Knight_Standard_Open_SO-100} equipped with a parallel gripper. All demonstration data were gathered exclusively via human teleoperation, and the system's control frequency was maintained at 30~Hz. Visual observations were acquired using two distinct camera setups: (1) an EMEET C960 web camera, rigidly mounted directly in front of the robotic arm to capture a third-person, frontal view of the workspace; and (2) an InnoMaker USB camera attached to the end-effector to provide a wrist-mounted perspective. All image streams were recorded at a resolution of $640 \times 480$ pixels.

To ensure task diversity, we systematically varied the initial placement of the manipulated objects during data collection. For the color-invariant task, we recorded 50 successful episodes featuring a single red ball placed across four distinct starting positions. For the color-dependent task, we recorded 240 successful episodes, capturing all possible spatial combinations of a red ball and a blue ball distributed across the four designated initial positions.

\section{Ablation Study}\label{sec:ablation}

\subsection{Evaluation on Objects with Similar Hues}
To further evaluate the color discrimination capability of ChromaGuard, we conduct an additional color-dependent task in which the robot must distinguish between red and pink objects (Fig.~\ref{fig:similar_hue}). Since red and pink share similar hues and differ mainly in saturation and value, both of which ChromaGuard perturbs during training, this task serves as a challenging test of fine-grained color discrimination. We follow the same protocol as the color-dependent task in \S\ref{sec:experimental_setup_real}, using SmolVLA, conducting a total of 40 trials.
As shown in Table~\ref{tab:similar_hue}, ChromaGuard achieves the highest SR in both benign and attacked scenarios (82.5\% and 72.5\%, respectively). In contrast, Naive-Aug degrades the benign SR from 65.0\% to 50.0\%, and 75.0\% of its failures are caused by grasping the incorrectly colored object, consistent with our findings in \S\ref{sec:color_tasks}. Under attack, the SR of the Baseline drops to 20.0\%, while ChromaGuard maintains 72.5\%. Note that all failures of ChromaGuard under attack stem from grasping the incorrectly colored object, indicating that distinguishing similar hues under targeted illumination remains challenging. Nevertheless, these results demonstrate that ChromaGuard retains fine-grained color discrimination even between objects with similar hues.

\subsection{Ablation on Perturbed Properties and Ranges}
We conduct an ablation study on which visual properties to perturb and how widely to perturb them, to validate the design choices of ChromaGuard, which fixes hue while perturbing value, sharpness, contrast, and saturation over wide ranges. To examine whether each of these perturbations is necessary, we evaluate variants in which one additional property (i.e., value, sharpness, contrast, or saturation) is fixed alongside hue. We also evaluate a narrow-range variant, which perturbs value and saturation within narrower ranges (see Appendix~\ref{sec:model_architecture} for exact values), to assess the impact of the perturbation range. All variants are evaluated on the color-dependent task using SmolVLA.
Table~\ref{tab:ablation_aug} shows the results. ChromaGuard achieves the highest SR under attack (88.8\%), while no variants improve robustness. Fixing value causes the most severe degradation, dropping the SR to 67.5\% and 60.0\% in benign and attacked scenarios, respectively. Fixing sharpness maintains a comparable benign SR (97.5\%), but its SR under attack falls to 62.5\%. The narrow-range variant is the least robust, achieving only 45.0\% SR under attack. Notably, none of the variants grasps an incorrectly colored object in benign scenarios. This suggests that preserving hue alone is sufficient to retain color discrimination, whereas wide perturbations on the remaining properties are key to robustness against spotlight attacks.

\end{document}